# Multi-Agent Multimodal Large Language Model Framework for Automated Interpretation of Fuel Efficiency Analytics in Public Transportation

Zhipeng Ma [1], Ali Rida Bahja [2], Andreas Burgdorf [2], André Pomp [2], Tobias Meisen [2], Bo Nørregaard Jørgensen [1,*] and Zheng Grace Ma [1]

[1] SDU Center for Energy Informatics, The Maersk Mc-Kinney Moller Institute, Faculty of Engineering, University of Southern Denmark, DK-5230 Odense, Denmark; zhma@mmmi.sdu.dk (Z.M.); zma@mmmi.sdu.dk (Z.G.M.)

[2] Institute for Technologies and Management of Digital Transformation, University of Wuppertal, D-42119 Wuppertal, Germany; bahja@uni-wuppertal.de (A.R.B.); burgdorf@uni-wuppertal.de (A.B.); pomp@uni-wuppertal.de (A.P.); meisen@uni-wuppertal.de (T.M.)

* Correspondence: bnj@mmmi.sdu.dk

**Abstract**

Enhancing fuel efficiency in public transportation requires the integration of complex multimodal data into interpretable, decision-relevant insights. However, traditional analytics and visualization methods often yield fragmented outputs that demand extensive human interpretation, limiting scalability and consistency. This study presents a multi-agent framework that leverages multimodal large language models (LLMs) to automate data narration and energy insight generation. The framework coordinates three specialized agents, including a data narration agent, an LLM-as-a-judge agent, and an optional human-in-the-loop evaluator, to iteratively transform analytical artifacts into coherent, stakeholder-oriented reports. The system is validated through a real-world case study on public bus transportation in Northern Jutland, Denmark, where fuel efficiency data from 4006 trips are analyzed using Gaussian Mixture Model clustering. Comparative experiments across five state-of-the-art LLMs and three prompting paradigms identify GPT-4.1 mini with Chain-of-Thought prompting as the optimal configuration, achieving 97.3% narrative accuracy while balancing interpretability and computational cost. The findings demonstrate that multi-agent orchestration significantly enhances factual precision, coherence, and scalability in LLM-based reporting. The proposed framework establishes a replicable and domain-adaptive methodology for AI-driven narrative generation and decision support in energy informatics.

## 1. Introduction

Scientific research and real-world operations are increasingly becoming data-intensive, involving large and heterogeneous datasets. Extracting meaningful insights from these data and communicating them effectively to diverse stakeholders remains a critical challenge [1]. Conventional analytical tools and methodologies often struggle to translate complex multimodal data into actionable knowledge due to limitations in interpretability, context-awareness, and integration capabilities [2,3].

In domains such as data science, business analytics, and urban mobility, professionals frequently produce reports that integrate text and visuals to communicate results clearly and coherently. These reports aim to present complex information in a clear, structured, and accessible manner [4,5]. Central to this practice is the concept of visual data storytelling, which combines charts, tables, and explanatory narratives into narrative formats that enhance comprehension and improve communication of analytical findings [6].

Recent advances in artificial intelligence, particularly multimodal Large Language Models (LLMs), offer promising capabilities for automating such storytelling tasks [7,8]. These models demonstrate robust competence in interpreting heterogeneous data sources, including textual, numerical, and visual inputs, and generating context-aware narratives [9–11]. For example, LLM4Vis [6] introduced a ChatGPT-based prompting technique capable of generating human-like explanations of charts and tables using GPT-3.5. Similarly, multimodal datasets like those in [11] have been used to fine-tune LLMs for enhanced performance in diagram interpretation.

Parallel developments in multi-agent systems have shown promise in orchestrating distributed analytical workflows. Recent frameworks have explored agent-based LLM architectures for mimicking human storytelling and verification processes. For example, [4] proposes a two-agent setup for data interpretation and narrative verification, while the Data Director framework [12] employs LLM agents to automate animated data video creation. Likewise, LLaMS [13] generates expressive multimodal stories using coordinated agents.

Despite these advances, current research largely focuses on isolated subtasks, such as image-text reasoning or chart summarization, without full integration into end-to-end energy analytics pipelines. Most frameworks do not leverage the global knowledge embedded in LLMs to generate comprehensive, data-supported recommendations, and rarely address stakeholder communication needs in energy-critical application domains such as transportation. Moreover, many applied analytics projects produce fragmented outputs (e.g., figures, tables, summaries), requiring time-consuming manual synthesis to produce decision-ready reports. This is an approach that is neither scalable nor consistent.

Within the emerging field of energy informatics, there is a growing need for systems that combine the interpretive capabilities of LLMs with domain-specific validation and transparency mechanisms. Current energy analytics pipelines often fail to deliver contextualized narratives that bridge analytical complexity and stakeholder needs. This limitation underscores a broader gap between the growing analytical sophistication of industrial data systems and the communicative, decision-oriented insights demanded by practitioners and policymakers.

To address these challenges, this study proposes a modular and scalable multi-agent framework that integrates multimodal LLMs, structured evaluation, and stakeholder-oriented reporting within an applied energy informatics setting. While LLMs are known to have limitations in symbolic and logical reasoning [14,15], they are well-suited to communicative and descriptive reasoning tasks, such as data storytelling, where they have shown notable effectiveness [3,4,16]. The framework introduces a layered reasoning architecture in which narration, validation, and synthesis are performed by specialized agents operating in an interconnected workflow. This configuration allows iterative narrative refinement, factual verification, and explainability, extending beyond traditional one-shot or single-agent approaches. By combining the narrative generation capabilities of multimodal LLMs with the coordination strengths of agent-based systems, the proposed framework bridges the gap between complex analytics and stakeholder decision support in energy-focused transportation planning.

The proposed architecture consists of three specialized agents:
- A data narration agent, which interprets multimodal inputs and generates textual narratives;
- An LLM-as-a-judge agent, which evaluates narrative quality, coherence, and relevance;
- An optional human-in-the-loop evaluator, which enables expert refinement in sensitive applications.

The framework operates across four stages: (i) raw data description, (ii) data modeling, (iii) post hoc analytics, and (iv) integration and narrative reporting. The final report generation is fully automated and formatted for non-technical stakeholders. The full system architecture is detailed in Section 3.

The scientific contribution of this work lies in demonstrating how foundation models can be systematically adapted and evaluated within real-world analytical environments, bridging the gap between AI-based storytelling and evidence-driven decision support. Methodologically, the framework formalizes an interpretable, verifiable, and reusable pipeline for automated data narration that ensures factual precision and transparency across reasoning stages. Empirically, the study provides a replicable evaluation protocol that quantifies narrative accuracy, informativeness, and computational efficiency across diverse LLM architectures. Applied within the context of fuel-efficiency analysis for public transportation in Northern Jutland, Denmark [17,18], the framework illustrates how AI-assisted narration can generate interpretable insights that support cost reduction and emission efficiency. These contributions establish a rigorous foundation for integrating intelligent language-based reasoning into applied energy analytics.

The remainder of this paper is structured as follows. Section 2 reviews relevant background and related work. Section 3 presents the proposed multi-agent framework in detail. Section 4 describes the case study and dataset. Section 5 reports the experimental setup and results. Section 6 discusses findings and outlines directions for future work, and Section 7 concludes the study.

## 2. Related Work

*2.1. Multimodal Large Language Models*

Multimodal LLMs represent a significant evolution in artificial intelligence by enabling the integration of multiple data modalities, including text, images, and numerical data, within a unified processing framework. These models aim to perform complex reasoning tasks that require a deep understanding of heterogeneous inputs, thereby extending the application scope of LLMs across scientific, industrial, and creative domains. A core task in the development of multimodal LLMs lies in establishing effective mechanisms to support collaborative inference and cross-modal reasoning [19]. One of the most active areas of research involves integrating visual comprehension into LLMs, allowing them to process and interpret charts, diagrams, and other scientific visual content. This capability has garnered growing interest from the research community, as it paves the way for broader and more versatile applications of LLMs in domains such as education, scientific communication, and industrial analytics [13].

Recent state-of-the-art models such as GPT-4.1 [20], Claude Opus 4 [21], and Gemini 2.5 Pro [22] have demonstrated strong performance in tasks like multimodal reasoning, visual question answering, and image-text alignment. These models benefit from advanced training techniques such as reinforcement learning with human feedback (RLHF), which enhances their ability to understand and relate multimodal content. However, the use of such high-performance commercial models often entails substantial cost, posing practical challenges for deployment. To address this, more cost-effective alternatives, such as GPT-4.1-mini [20], o4-mini [23], Claude 3.5 Haiku [24], and Gemini 2.5 Flash [25], have been developed. Therefore, selecting the appropriate model for a given application requires careful evaluation of the trade-offs between cost, computational efficiency, and inference accuracy during framework development.

*2.2. Prompt Engineering*

Prompt engineering has emerged as a critical technique for enhancing the performance and adaptability of pre-trained LLMs. By crafting precise and contextually rich instructions, prompt engineering enables fine control over model outputs without requiring additional model training, thereby improving their effectiveness across a wide range of tasks and domains [26]. The significance of prompt engineering is underscored by its capacity to steer model responses, enhancing the adaptability and applicability of LLMs across diverse sectors.

Standard prompting strategies primarily include zero-shot and few-shot prompting. In zero-shot prompting [27], models are guided to perform tasks without any prior examples, relying solely on the prompt's phrasing. Few-shot prompting [28], by contrast, includes a small number of input-output examples within the prompt to help the model infer the desired task behavior. Even a few high-quality examples can substantially improve performance on complex or ambiguous tasks.

However, these standard strategies often fall short in scenarios that demand multi-step or structured reasoning. To address this limitation, Chain-of-Thought (CoT) prompting [29] has been introduced to enable LLMs to produce intermediate reasoning steps before arriving at a final answer. This method significantly improves logical coherence and interpretability in model outputs. Building on CoT, Chain-of-Table prompting [16] has been built for step-by-step tabular reasoning, where the model dynamically generates and executes SQL or data frame operations. To further enhance reasoning robustness, the Contrastive Chain-of-Thought (CCoT) prompting [30] has been introduced, which provides both valid and flawed reasoning demonstrations. This contrastive learning framework enables models to distinguish between correct and incorrect logic, navigating a map with clearly marked right and wrong paths, ultimately fostering more reliable and explainable outputs.

Given the diversity of task types and data modalities involved in our framework, selecting an appropriate prompt strategy is essential to ensure both performance and interpretability.

*2.3. Multimodal LLMs for Data Narration*

Automated story generation is an open-ended task focused on producing coherent sequences of events based on predefined criteria or input data [31]. Recent advances in multimodal LLMs have significantly improved performance in this domain. These models are capable of generating fluent, contextually appropriate narratives by iteratively incorporating information from both the narrative plan and the evolving state of the story into the prompt. Empirical studies [2,4] have confirmed the effectiveness of LLMs in producing short, coherent, and natural-sounding stories, highlighting their potential for applications in dynamic content generation.

Related to narrative generation, recent research [11,32,33] has explored the ability of multimodal LLMs to interpret complex data-rich visuals, such as scientific charts, tables, and documents. Notably, studies focusing on understanding visualizations within academic literature demonstrate the potential of these models to generate insightful narratives from structured and semi-structured data, enabling more scalable and nuanced approaches to scientific data analysis.

Complementing these developments, agent-based systems [2,4] have emerged as a promising paradigm for automating narrative workflows. For instance, DS-Agent [34] integrates LLMs with case-based reasoning to automate data science tasks, significantly reducing manual intervention. Building on this concept, the study in [12] has introduced a multi-agent framework leveraging LLMs to generate animated data videos, further expanding the boundaries of automated data storytelling.

While recent work has advanced data narration using LLMs, most approaches rely on fine-tuning models for specific domains, limiting their generalizability. Few studies explore the zero-shot capabilities of LLMs to leverage their global knowledge for narrative generation across diverse industrial contexts. Furthermore, existing research tends to focus on descriptive outputs rather than generating actionable recommendations tailored to stakeholder needs. In cases such as improving fuel efficiency for public transportation, these limitations are significant: decision support requires both interpretable narratives and practical suggestions. This highlights the need for frameworks that combine zero-shot narration, automated evaluation, and stakeholder-aware recommendation generation toward integrated, real-world decision support.

## 3. Materials and Methods

The proposed intelligent multi-agent framework integrates multimodal LLMs to automate the generation of scientific narratives and actionable industrial insights from industrial data science projects. This section introduces the detailed system architecture in Section 3.1 and the functional workflow in Section 3.2.

### 3.1. Multi-Agent Architecture

The objective of this study is to develop a novel multi-agent-based approach capable of generating coherent data story narration from scientific data analytics and producing actionable recommendations for stakeholders. At its core, this approach distributes complex reasoning and evaluation tasks across specialized agents, enabling iterative and intelligent refinement. To ensure the accuracy and clarity of the generated narratives, a multi-agent architecture is designed, referred to as a block. This architecture is illustrated in Figure 1.

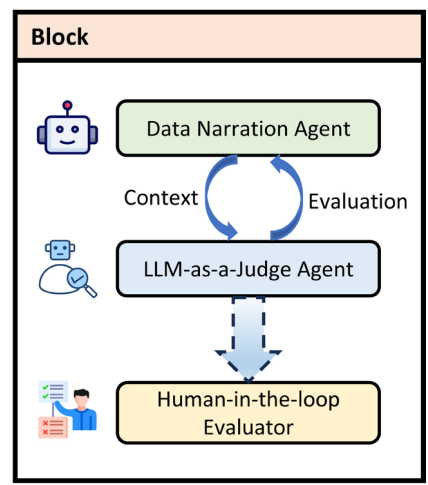

**Figure 1.** The multi-agent architecture for data narration.

This architecture comprises three specialized agents, each fulfilling a distinct role within the data-to-insight workflow. The process begins with the data narration agent, which serves as the initial storyteller. Its role is to bridge the gap between raw multimodal inputs, such as charts, tables, and numerical results, and human-understandable narratives, leveraging the LLM's global knowledge to generate context-aware textual outputs. Next, the LLM-as-a-judge agent acts as an automated reviewer, providing critical reasoning and iterative evaluation to assess narrative quality, internal consistency, and alignment with analytical objectives. Finally, the human-in-the-loop evaluator introduces an expert oversight, refining and validating machine-generated outputs in domain-sensitive applications if necessary. Together, these agents create a dynamic workflow that advances from data interpretation to actionable insights, supporting real-world decision-making processes.

The data narration agent is responsible for processing multimodal inputs such as charts, tables, and numerical results, and transforming them into coherent, context-aware textual narratives with an LLM. The LLM-as-a-judge agent then assesses the generated narratives for quality, internal consistency, and alignment with the intended analytical objectives. To enhance the reliability and domain relevance of the outputs, an optional human-in-the-loop evaluator is included. This agent provides expert oversight and enables the validation or refinement of the machine-generated narratives in applications that require domain-specific sensitivity.

Within the block, the data narration agent is automatically triggered upon a dataset upload by a user. It then interprets multimodal inputs derived from data analytics tasks, generating corresponding textual explanations and descriptive narratives. These outputs are then passed to the LLM-as-a-judge agent, which systematically assesses the generated narratives across four key dimensions: clarity, relevance, insightfulness, and contextualization. These dimensions are derived from a review of the literature, as discussed in [2,4,35]. Each evaluation dimension is scored on a scale from 0 to 4, with higher scores indicating better performance. To ensure reliability, the agent's stability is assessed by running repeated evaluations on identical narratives. When setting the hyperparameter "temperature = 0.1", the agent consistently produced identical scores and comments across runs, confirming its deterministic behavior and robustness. The details of the defined scale are introduced in Listing A4 in Appendix A. This evaluation basis, including the definitions of each dimension and the 0–4 scoring criteria, is also provided to the LLM-as-a-judge agent via prompt instructions to ensure consistent and aligned assessments. Based on these scores, a feedback report is generated and returned to the data narration agent for validation and potential refinement of the data story. This iterative process continues until a predefined stopping condition is met, such as achieving a target average score or reaching a maximum number of refinement cycles.

Next, a human-in-the-loop evaluator can be engaged when a domain expert deems intervention necessary, such as in cases involving complex inputs or where additional contextualization and background information are required. In such instances, the expert reviews the generated narrative and issues explicit validation commands, providing judgments on its accuracy, interpretability, and overall utility. This input supports further refinement of the narrative and helps ensure its alignment with domain-specific expectations.

*3.2. Workflow of the Multi-Agent Framework*

To transform analytic artifacts from industrial data science projects, such as model outputs, charts, and tables, into structured, context-rich reports that support industrial decision-making, this study develops a structured multi-agent framework in this section. The framework distributes analytical interpretation across modular stages, where each stage builds upon the cumulative knowledge and narratives of preceding stages. This design enables iterative refinement through the block designed in Section 3.1, and supports seamless integration of machine-generated insights into human-centered decision-making processes. By maintaining interconnected streams of contextual background and narrative outputs, the system ensures transparency and traceability throughout the workflow. The overall architecture systematically converts multimodal artifacts into coherent, context-rich reports aligned with the needs of diverse industrial stakeholders. The overall workflow of the framework is illustrated in Figure 2.

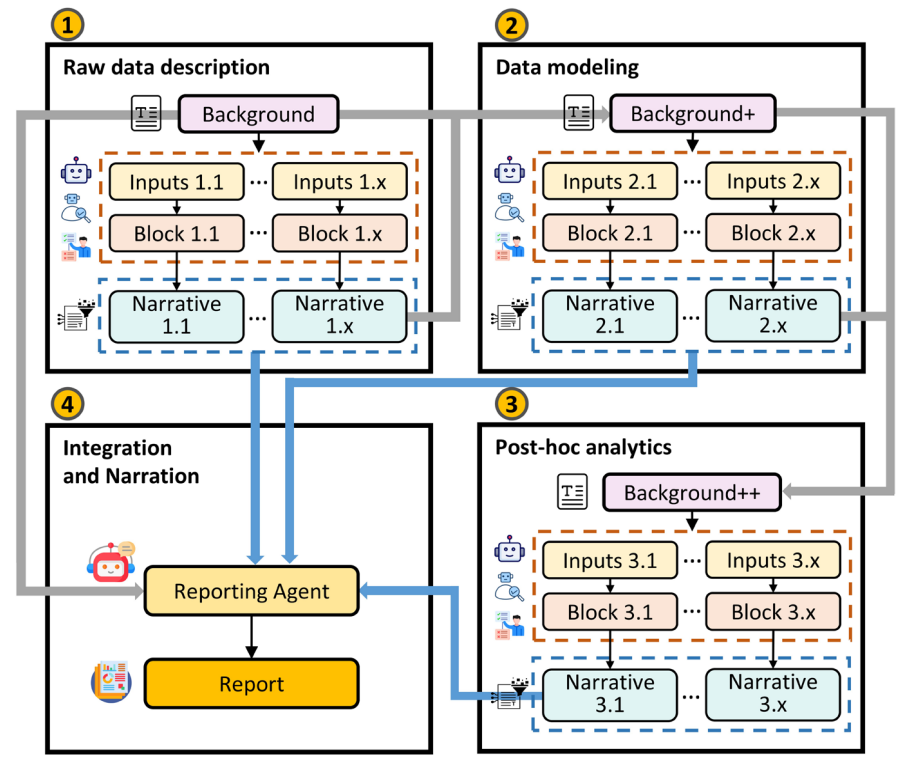

**Figure 2.** The workflow of the proposed LLM-agent framework for data narration and reporting.

As illustrated in Figure 2, the framework proceeds through four connected stages: raw-data description, data modeling, post hoc analytics, and integration and narration. The contextual knowledge in each subsequent stage builds cumulatively, incorporating both the background from the preceding stages and the narrative outputs generated at those stages. For instance, the "Background+" in Stage 2 integrates the "Background" in Stage 1 along with Narrative 1.1–1.x.

Each of the first three stages also includes a series of modular processing units. The yellow boxes labeled "Input x.y" denote the multimodal data elements provided for interpretation, such as figures, charts, or tables. These are processed by the orange boxes labeled "Block x.y", which correspond to the multi-agent architecture outlined in Figure 1. Each block takes a specific input to produce a textual output, represented by the green boxes labeled "Narrative x.y". For example, "Inputs 1.1" consists of a combination of figures and tables, then "Block 1.1" processes these using the multi-agent approach in Figure 1, yielding Narrative 1.1, a natural language explanation summarizing trends or insights drawn from "Inputs 1.1".

These stages are linked by two distinct information streams. One stream, represented in grey, accumulates and transmits contextual background knowledge across stages: Stage 2 receives enriched context from Stage 1, Stage 3 builds on Stage 2, and Stage 4 directly reuses the background from Stage 1 to ensure fidelity in the final reporting. The second stream, represented in blue, collects and transmits narrative outputs from Stages 1–3, including "Narrative 1.1–3.x", directly to the reporting agent in Stage 4.

The workflow begins with the description of the raw data used in the project. The users, such as data scientists or end-users, provide multimodal raw data descriptions. The objective at this stage is to characterize the structure, distribution, and quality of the input data. Therefore, the system receives the initial project description, referred to as "Background", and processes a set of multimodal artifacts, denoted as "Input 1.1–1.x". For instance, if the project involves two datasets, "Input 1.1" may include data distribution charts and statistical tables for dataset 1, while "Input 1.2" pertains to those for dataset 2. Each input component is processed in parallel and, respectively, by "Block 1.1–1.x" introduced in Figure 1. These blocks generate narratives, referred to as "Narrative 1.1–1.x", respectively, that summarize the key characteristics of the data. To ensure consistency and coherence, the narratives are produced using a predefined general template, developed based on insights from the literature [4,16,30] and the needs of end-users. The template is demonstrated in Listing A5 in Appendix A. The resulting narratives "Narrative 1.1–1.x" are preserved in the narrative stream, while the combination of the original project description "Background" and the generated descriptions are compiled into an enriched contextual file, referred to as "Background+". This enhanced background is then forwarded to support the subsequent stage of the workflow.

The second stage, data modeling, centers on the modeling aspect of data analytics. The objective of this stage is to explain the analytical results using statistical or machine learning models. The enriched background, "Background+", provides the necessary context for each multimodal modeling artifact, denoted as "Input 2.1–2.x". For example, if the project employs two analytical methods, such as clustering and causal inference, then "Input 2.1" may consist of charts, figures, or tables showcasing clustering results, while "Input 2.2" corresponds to the outputs of the causal inference analysis. Each input component is processed in parallel and independently by modules referred to as "Block 2.1–2.x". These blocks generate narratives, "Narrative 2.1–2.x", which summarize and interpret the modeling results using the same general template introduced in Listing A5 The newly generated narratives are appended to the existing narrative stream. Additionally, a further enriched contextual file, denoted as "Background++", is created by combining "Background+" with the modeling narratives to support Stage 3.

In the third stage, the system explains post hoc analytics to further interpret model outputs. Leveraging the comprehensive background "Background++" that now includes the project context, data summaries, and modeling explanations, the system processes a range of multimodal post hoc artifacts. These may include SHAP value distributions, counterfactual explanations, and statistical evaluations of model residuals. For instance, "Input 3.1" might consist of multimodal representations of SHAP value distributions, "Input 3.2" of counterfactual explanations, and "Input 3.3" of residual diagnostics. Each input component is handled in parallel by corresponding modules, labeled "Block 3.1–3.x", which generate explanatory narratives, "Narrative 3.1–3.x", according to the template introduced in Listing A5. The system itself does not autonomously select or suggest post hoc analysis methods; it relies entirely on artifacts provided by data scientists or domain experts. Following this stage, all relevant contextual information is consolidated and ready for the final stage, Integration and Narration.

The final stage of the workflow integrates all previously generated narratives, "Narrative 3.1–3.x", and the background context, "Background", into a cohesive technical report. A dedicated agent, denoted as the reporting agent, receives the complete set of narratives "Narrative 3.1–3.x" along with the original project description "Background". Without modifying the factual content, this agent standardizes terminology, eliminates redundancies, and organizes the material into a logically structured and coherent document. The final output report includes an executive summary, a methodological exposition, analytical findings, and actionable recommendations derived from the multimodal scientific data narratives. The report is formatted to facilitate interpretation by industrial stakeholders with diverse levels of technical expertise. The level of technical detail and complexity can be adjusted by modifying the corresponding prompts.

Overall, the proposed framework offers a principled solution for transforming complex and multimodal analytical results into high-quality documentation. By maintaining a strict separation between cumulative context and interpretive content, supporting parallel processing through modular blocks, and preserving an auditable flow of information across all stages, the system enables transparent, reliable, and stakeholder-aligned reporting in data-intensive industrial environments.

It is important to note that the number of narratives generated in each stage depends entirely on the multimodal artifacts produced by the data analytics project. The system processes each artifact as input and does not intervene in determining their quantity or content. Hyperparameters such as the temperature and token limits are configured to ensure consistent narrative generation but do not influence the number of narratives. To illustrate the application of this workflow and provide a concrete example, Section 4 presents a case study demonstrating the system in an industrial data analytics context.

## 4. Case Study

The proposed framework is applied to an industrial data analytics project that utilizes GMM clustering to analyze the fuel efficiency data of bus trips from the public transportation system in North Jutland, Denmark [17]. In this context, a bus trip is defined as a complete journey from the first stop to the destination according to a scheduled route, occurring within a specific time window. The dataset contains a total of 4006 recorded trips. The key feature under analysis, fuel efficiency, represents fuel consumption in liters per 100 km (L/100 km), following the European Union standard.

The project produces a total of eleven scientific charts to represent different stages of the analysis. In Stage 1, the raw data is described using a histogram that illustrates the distribution of fuel efficiency; this visualization is referred to as "Input 1.1" in Figure 2. Directly providing extensive raw datasets to LLM often results in the generation of fabricated or hallucinated content. To mitigate this issue, data distribution figures are used, which more effectively convey the key characteristics of the raw data and reduce the risk of misinformation during narrative generation. In Stage 2, clustering results are visualized through a combined figure featuring a probability density estimate (PDE) distribution and a scatter plot of the clustered data, serving as "Input 2.1". The post hoc analysis, corresponding to Stage 3, is organized into three thematic groups. The first group, "Input 3.1", focuses on the distribution of drivers across the four clusters and employs a stacked bar chart alongside a histogram to display entropy distribution. The second group, "Input 3.2", examines the distribution of routes across the clusters, using similar visualization methods. The third group, "Input 3.3", analyzes the distribution of route types, including urban, rural, and highway, within each cluster. This group is visualized through a ternary scatter plot, complemented by four histograms that illustrate the entropy distribution of route types within each cluster. This group is represented by a ternary scatter plot and four corresponding histograms showing the entropy distribution of route types under each cluster.

## 5. Experiments and Results

*5.1. Experiment Design and Settings*

A series of experiments is conducted to evaluate the effectiveness of the proposed framework and validate its components.

The first set of experiments focuses on assessing the performance of different prompting strategies and LLMs in the data narration task. Three prompting strategies are selected as candidates: CoT [29], CCoT [30], and DataNarrative (DN) [4]. CoT is selected as a widely used prompting technique and a standard benchmark for eliciting step-by-step reasoning in LLMs. CCoT extends CoT by first generating structured contextual representations, making it suitable for multimodal reasoning tasks. DN is the most related published approach to this study. All three approaches are implemented in a zero-shot setting, chosen for their speed and efficiency in real-world applications.

In zero-shot scenarios, where no task-specific examples are provided, prompting strategies can enhance LLM performance by guiding inference. The CoT uses simple cues like "Let's think step by step" in zero-shot context to elicit multi-step reasoning [29]. The CCoT first generates a structured scene graph from an image, then reasons over it alongside the original prompt, enabling compositional understanding without fine-tuning or annotations [30]. The DataNarrative prompting operates on structured inputs such as tables and charts, prompting the model to produce coherent narratives that highlight trends, anomalies, and relationships directly from the raw data [4].

Five LLMs are evaluated, including GPT-4.1 mini [20], o4-mini [23], Claude 3.5 Haiku [24], Gemini 2.5 flash [25], and LLama 4 maverick [36]. These models represent the latest lightweight versions developed by leading LLM providers. They offer a practical balance between performance and cost, with LLama 4 Maverick being open-source and freely available for public use. The experimental design evaluates all 15 combinations of the selected prompting strategies and LLMs. Performance is assessed using five criteria:

- Execution time, measured in seconds, quantifies the time required to generate each narrative;
- Cost, measured in cents (0.01 USD), reflects the expense of using the model based on its API pricing;
- Narrative length, defined as the total word count of the generated output;
- Informativeness, measured as the number of distinct information points extracted from the visual input;
- Accuracy, calculated as the percentage of correct information points relative to the total number of points generated.
- Accurate Narrative Information Density (ANID), defined as the number of correct distinct information points per 100 words in a generated narrative.

Especially in this study, an information point is defined as a single word or short phrase that conveys one discrete piece of meaning relevant to the content of the input being narrated. For example, in the sentence "The $x$-axis denotes fuel efficiency measured in liters per 100 km", the term "fuel efficiency" forms one information point by identifying the variable on the $x$-axis, whereas "liters per 100 km" forms a second information point by specifying that variable's unit. The correctness of each information point is assessed by cross-checking it against domain-specific ground truth, common-sense knowledge, and the content of the original inputs.

Together, these metrics provide a comprehensive basis for evaluating the efficiency, precision, and overall quality of prompt–LLM configurations in the context of automated data narration.

The second set of experiments focuses on selecting the most suitable LLM for Stage 4 of the framework, which is responsible for generating the final report. Based on the results of the first experiment demonstrated in Table 1. In Section 5.2, which indicates that OpenAI models outperform others in data narration, three OpenAI models are selected as candidates: GPT-4.1 mini, GPT-4.1, and o4-mini. The quality and utility of the generated reports are evaluated using the following criteria:

- Flesch Reading Ease (FRE) [37]: A readability score that indicates how easy a text is to read, with higher values corresponding to easier reading (typically ranging from 0 to 100).

- Flesch-Kincaid Grade Level (FKGL) [37]: A metric that estimates the U.S. school grade level required to understand the text, with lower values indicating greater accessibility.
- Automated Readability Index (ARI) [38]: Another readability measure based on word and sentence length, providing a numeric grade level; lower values reflect simpler language.
- Cost: The monetary cost of generating each report, measured in cents (USD 0.01), based on API pricing for each model.
- Report length (RL): The total word count of the generated report, reflecting the amount of explanatory content produced.
- Number of recommendations with data support (NRDS): The subset of recommendations explicitly grounded in the data narratives.
- Percentage of recommendations with data support (PRDS): The percentage of recommendations explicitly grounded in the data narratives relative to the total number of actionable suggestions included in the report.

Together, these metrics offer a comprehensive basis for comparing the readability, efficiency, and practical utility of each model in producing stakeholder-ready reports.

The third set of experiments consists of ablation studies designed to evaluate the contribution of each agent in the framework to the quality of the final report. Specifically, the study examines the effectiveness of the background description, the LLM-as-a-judge agent, and the human-in-the-loop evaluator by analyzing the impact of their removal. The experiments assess how each agent influences key aspects of the data narrative, including coherence, informativeness, and reliability.

The results of the first two experimental sets are presented in Section 5.2, while the outcomes of the ablation studies are detailed in Section 5.3. The Representative examples of data narration are provided in Tables A1 and A2 in Appendix B. The key prompts and their associated contextual inputs are illustrated in Appendix A. The final generated report is shown in Listing A6 in Appendix C.

### 5.2. Prompts and LLM Selection

This section evaluates the performance of both proprietary and open-source LLMs along with multiple prompting strategies, based on the experimental design described in Section 5.1.

Table 1 presents the evaluation metrics and corresponding scores for various combinations of LLMs and prompting strategies applied to scientific chart narration. The values preceding the "±" symbol indicate the evaluation scores introduced in Section 5.1, while the values following it denote the standard errors of repetitions. Llama 4 Maverick does not have an associated cost, as it is open source and available free of charge, and is therefore marked as "N/A." In the context of method selection, informativeness and accuracy are prioritized, as they directly reflect the amount and accuracy of information conveyed. Execution time and cost are considered secondary but remain important for practical deployment.

**Table 1.** Evaluation metrics for different LLM and prompt strategy combinations applied to scientific chart narration.

| LLM | Prompting | Time (s) | Cost (cent) | Narrative length | Informativeness | Accuracy (%) | ANID |
|---|---|---|---|---|---|---|---|
| GPT-4.1 mini | CoT | 13.27 ± 0.849 | 0.24 ± 0.008 | 305.4 ± 10.54 | 20.91 ± 0.567 | 97.31 ± 1.295 | 6.73 ± 0.267 |
| | CCoT | 41.48 ± 2.383 | 0.62 ± 0.020 | 288.2 ± 13.14 | 20.36 ± 0.646 | 98.52 ± 1.056 | 7.13 ± 0.384 |
| | DN | 10.91 ± 0.411 | 0.23 ± 0.007 | 315.9 ± 10.83 | 25.55 ± 0.999 | 95.45 ± 1.518 | 7.75 ± 0.273 |
| o4-mini | CoT | 14.53 ± 0.595 | 0.96 ± 0.023 | 285.1 ± 11.37 | 21.91 ± 0.698 | 96.97 ± 1.158 | 7.55 ± 0.334 |
| | CCoT | 39.17 ± 1.823 | 2.37 ± 0.056 | 257.6 ± 11.92 | 20.45 ± 0.754 | 96.17 ± 1.776 | 7.76 ± 0.397 |
| | DN | 19.17 ± 1.301 | 1.10 ± 0.038 | 308.9 ± 17.42 | 25.45 ± 1.474 | 94.43 ± 1.830 | 7.80 ± 0.232 |
| Claude 3.5 Haiku | CoT | 9.80 ± 0.417 | 0.42 ± 0.002 | 207.3 ± 5.20 | 16.70 ± 0.736 | 84.46 ± 4.597 | 6.94 ± 0.539 |
| | CCoT | 21.69 ± 0.617 | 1.05 ± 0.040 | 192.2 ± 4.81 | 17.10 ± 0.767 | 91.14 ± 4.895 | 8.35 ± 0.700 |
| | DN | 12.65 ± 1.667 | 0.45 ± 0.002 | 230.8 ± 3.97 | 19.60 ± 0.710 | 91.48 ± 3.014 | 7.84 ± 0.475 |
| Gemini 2.5 Flash | CoT | 8.96 ± 0.623 | 0.11 ± 0.007 | 287.5 ± 11.14 | 20.55 ± 0.965 | 92.91 ± 0.018 | 6.67 ± 0.300 |
| | CCoT | 22.87 ± 1.297 | 0.29 ± 0.007 | 227.2 ± 8.30 | 16.82 ± 6.533 | 90.16 ± 3.404 | 6.82 ± 0.469 |
| | DN | 5.14 ± 0.249 | 0.07 ± 0.003 | 278.9 ± 16.49 | 21.27 ± 1.554 | 90.60 ± 3.011 | 7.19 ± 0.304 |
| Llama 4 Maverick | CoT | 26.00 ± 2.615 | N/A | 220.1 ± 13.41 | 18.64 ± 0.412 | 88.64 ± 2.756 | 7.71 ± 0.398 |
| | CCoT | 56.17 ± 4.459 | N/A | 228.7 ± 7.79 | 17.91 ± 0.537 | 91.36 ± 2.938 | 7.19 ± 0.304 |
| | DN | 27.24 ± 2.387 | N/A | 242.4 ± 11.57 | 19.18 ± 0.861 | 89.83 ± 2.409 | 7.13 ± 0.247 |

For LLM selection, the LLMs are first evaluated based on accuracy, ANID, narrative length, and informativeness. Among the candidates, GPT-4.1 mini and o4-mini consistently outperform the others, especially in accuracy and informativeness, indicating that OpenAI's models are particularly effective for this task. While o4-mini incurs the highest cost, it remains within an acceptable range. GPT-4.1 mini, in contrast, is more cost-efficient and produces narratives that are both longer and more accurate. Although o4-mini achieves a higher ANID, indicating greater conciseness, it also generates a higher proportion of incorrect information. Given its superior balance between performance and cost, GPT-4.1 mini is selected for the subsequent experiments.

The prompting strategies are then compared. While CCoT achieves the highest accuracy and produces the most precise narratives, the improvement over CoT is relatively marginal. Notably, this modest gain requires a substantially higher computational overhead, including approximately three times the execution time and 2.5 times the cost, as well as resulting in shorter narratives and fewer extracted information points. DN generates the most extensive and concise descriptions but exhibits the lowest accuracy. In contrast, CoT demonstrates the most balanced performance across all evaluation dimensions, delivering competitive accuracy, more informative narratives, and significantly lower computational demands. Based on this comprehensive assessment, CoT is identified as the most effective and efficient prompting strategy for the task.

To ensure comparability with prior multimodal interpretation frameworks such as DS-Agent [30] and LLM4Vis [6], DN is included as a baseline, reflecting the narrative synthesis paradigm used in earlier single-agent systems. As shown in Table 1, GPT-4.1 mini + CoT achieves 97.3% accuracy compared with 95.4% for DN, along with lower cost and higher information density. These findings indicate that the proposed multi-agent orchestration and iterative evaluation substantially improve interpretive reliability relative to other narration systems.

Table 2 presents the evaluation metrics for three LLMs, including GPT-4.1, GPT-4.1 mini, and o4-mini, applied to the task of report generation. The metrics include readability scores (FRE, FKGL, and ARI), computational cost (in USD cents), report length (RL, in words), number of generated recommendations with data support (NRDS), and percentage of recommendations with data support (PRDS) as introduced in Section 5.1.

Table 2. Evaluation metrics for different LLMs applied to report generation.

| LLM | FRE | FKGL | ARI | Cost | RL | NRDS | PRDS |
|---|---|---|---|---|---|---|---|
| GPT-4.1 | 21.26 | 14.88 | 16.64 | 2.52 | 1080 | 5 | 100 |
| GPT-4.1 mini | 19.45 | 14.63 | 16.37 | 0.47 | 1127 | 5 | 100 |
| o4-mini | 28.50 | 12.95 | 14.77 | 1.52 | 982 | 3 | 75 |

Among these metrics, the most important are NRDS and PRDS, as they indicate the model's ability to generate actionable insights that are grounded in the analytical results. Both GPT-4.1 and GPT-4.1 mini achieve the highest values in these categories, each generating five recommendations, all of which are directly supported by the chart narratives. Although o4-mini performs best in terms of readability, achieving the highest FRE, the lowest FKGL and ARI, its lower number of supported recommendations and higher cost relative to GPT-4.1 mini make it less favorable overall. GPT-4.1 provides equally strong content quality but at a much higher cost compared to GPT-4.1 mini. Considering all factors, GPT-4.1 mini offers the best balance between effectiveness and efficiency. It matches the top performance in NRDS and PRDS, maintains acceptable readability, and has the lowest cost among the three models. Therefore, GPT-4.1 mini is selected as the most suitable model for report generation in this framework.

*5.3. Ablation Study*

To systematically evaluate the contributions of the individual components within the proposed multi-agent architecture presented in Figure 1, an ablation study is conducted. This study aims to discern the relative impacts of the CoT prompting approach, contextual background propagation (B), and the LLM-as-a-judge agent (E), alongside optional human oversight, on the quality of generated scientific narratives. Four distinct configurations were analyzed: CoT-only, CoT combined with contextual background (CoT+B), CoT with contextual background and the LLM-as-a-judge agent (CoT+B+E), and the full baseline comprising CoT, background, LLM-as-a-judge, and human-in-the-loop evaluator as detailed in Figure 1. As in the baseline experiments, all cases pass the LLM-as-a-judge evaluation without requiring validation. To demonstrate the functionality of the LLM-as-a-judge agent, the CoT+B+E experiment is designed to explicitly invoke the agent by forcing it to revise the narrative context once.

Each configuration is rigorously tested on the same set of eleven chart narration scenarios in Section 5.2, employing three primary metrics: narrative length, informativeness, and accuracy, as detailed previously in Section 5.1. Figure 3 illustrates these results, depicting the comparative performance across these metrics for all configurations tested.

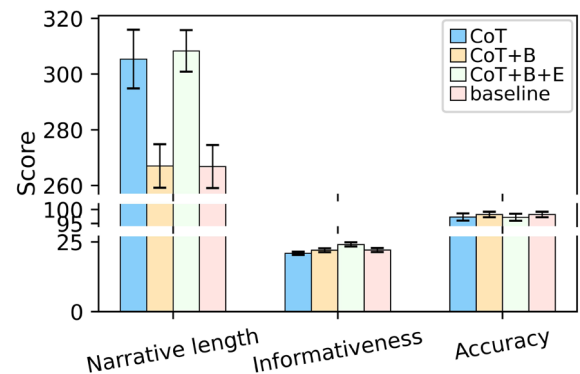

**Figure 3.** The evaluation scores of ablation studies. The x-axis indicates the evaluation criteria, the y-axis reflects the corresponding scores, and the legend visualizes corresponding configurations. The short black lines and the white gaps on the y-axis indicate axis breaks, which are used to conserve space and enable clearer visual comparison.

The results illustrated in Figure 3 suggest that the baseline method demonstrates the most effective overall performance across key evaluation dimensions. In the accuracy category, the baseline outperforms all other prompting strategies, indicating that it generates the most correct and reliable outputs. While CoT+B exhibits performance that appears comparable across three evaluation criteria, this similarity is superficial. In the baseline, the E stage does not perform additional validation because all narratives successfully pass automated evaluation. Crucially, 10 out of the 11 narratives in the baseline require no human intervention, underscoring its inherent robustness and reduced dependency on downstream adjustments. It makes the overall procedure nearly identical to that of CoT+B. The single case that involves human intervention in the baseline leads to an improvement in narrative quality. This highlights its strength in producing factually accurate and relevant content, which is critical in tasks requiring precision.

Although the narrative length for the baseline is the shortest among the four methods, this should not be interpreted as a limitation. On the contrary, the shorter narrative length reflects the baseline's efficiency in communication, achieving similar levels of informativeness with significantly fewer words. This suggests that the baseline is capable of delivering concise yet meaningful responses, avoiding redundancy or unnecessary elaboration that may be present in longer narratives generated by other methods.

Furthermore, the informativeness score of the baseline remains competitive. Despite its brevity, it successfully conveys the essential information needed to understand the data, comparable to or slightly below that of the CoT+B+E method. This reinforces the idea that longer responses do not necessarily translate to more informative content. Instead, the baseline achieves a more optimal balance between clarity, informativeness, and brevity.

In contrast, while methods such as CoT and CoT+B+E produce longer narratives, their higher word counts do not correspond to improved accuracy. Their performance in accuracy falls behind the baseline. Although CoT+B+E achieves a slightly higher informativeness score compared to the baseline, this is likely due to the inclusion of more contextual and descriptive details. However, this added detail introduces a greater risk of irrelevant or inaccurate statements, which ultimately undermines factual precision. The results indicate that unnecessary validation in LLM-based data narration tasks may introduce additional content that is either irrelevant or inaccurate, ultimately reducing the overall factual precision of the narrative.

In summary, the baseline configuration proves to be the most effective approach overall, achieving the highest accuracy while maintaining strong informativeness and generating the most concise narratives. These results underscore its capacity to produce high-quality, precise, and communication-efficient outputs. The framework is intentionally designed to minimize, though not entirely eliminate, human intervention. While the vast majority of narratives are generated autonomously, selective expert input is incorporated in exceptional cases where refinement or domain-specific correction is required. This hybrid setup strikes a deliberate balance between automation and human oversight, supporting both the scalability and reliability of the system in real-world data analytics workflows.

## 6. Discussion

*6.1. Interpretation of Results and Integration with Prior Work*

The results from the case study and experimental evaluations demonstrate that the proposed multi-agent framework provides a robust solution for transforming complex energy analytics into interpretable and actionable narratives. The framework's ability to consistently produce informative and contextually accurate narratives from multimodal data, specifically in the context of public transportation fuel efficiency, marks a significant advancement over prior efforts in automated data storytelling.

Earlier studies, such as DataNarrative [4] and LLM4Vis [6], have shown that LLMs can describe individual charts or provide explanations of structured data. However, these efforts largely focus on visualization in isolation and do not account for the broader analytical pipeline, including model outputs, post hoc analysis, and full-report generation. In contrast, the present framework offers an integrated, multi-stage process that reflects the complexity of real-world energy informatics workflows. It supports iterative improvement, cross-agent validation, and modular reuse.

Additionally, previous agent-based LLM systems, such as LLaMS [13] and Data Director [12], have primarily been evaluated in synthetic or generalized storytelling tasks. By grounding the system in a real-world domain with operational relevance in public transport fuel efficiency, the present study extends the scope of LLM-based storytelling into applied energy analytics. The case study results show that GPT-4.1 mini with Chain-of-Thought prompting provides a good trade-off between accuracy, informativeness, and computational cost, consistent with findings in [29,30] on the effectiveness of prompt engineering in guiding LLM reasoning.

The ablation study offers empirical validation for the framework's agentic design. The inclusion of an LLM-as-a-judge agent contributed to measurable improvements in narrative coherence and factual accuracy, confirming the emerging role of LLM evaluators in automated reasoning workflows [7]. The optional human-in-the-loop component added further reliability, especially in contexts requiring expert judgment.

*6.2. Scientific Contributions and Novelty*

This work contributes a modular and scalable framework that unifies multimodal LLMs and multi-agent architectures within an applied energy informatics setting. Methodologically, the framework introduces a layered reasoning design where narration, validation, and synthesis are handled by specialized agents operating across an interconnected workflow. This architecture advances beyond prior one-shot or single-agent systems by enabling iterative narrative refinement and quality assurance within an automated process.

Scientifically, the study offers a replicable methodology for evaluating prompt-LLM configurations in complex, real-world analytics settings. It demonstrates that decision support systems powered by foundation models can move beyond static chart summarization into dynamic, context-aware reporting pipelines. The work also bridges the literature gap between AI-based storytelling and stakeholder-centered energy decision support, an intersection previously underexplored in both energy informatics and applied AI research.

Compared with existing multimodal frameworks such as LLM4Vis [6], DataNarrative [4], and DS-Agent [30], the proposed multi-agent system advances the state of the art by integrating multimodal understanding, iterative evaluation, and stakeholder-oriented reporting in a unified architecture. While LLM4Vis focuses on visualization selection and DS-Agent automates data-science workflows, the proposed framework extends these approaches through domain-adaptive, decision-focused narration supported by explicit agent coordination and factual verification. Quantitative results in Section 5 show that the multi-agent configuration consistently outperforms the DataNarrative baseline in accuracy, informativeness, and narrative consistency while maintaining comparable computational cost. Overall, the framework establishes a reproducible, interpretable, and scalable foundation for AI-assisted insight generation in energy and industrial systems.

*6.3. Methodological Limitations*

Several methodological limitations should be acknowledged. First, system performance depends on the pretraining scope and domain coverage of the underlying language models. Although GPT-4.1 mini performs well in general contexts, its lack of fine-tuning on energy-specific corpora can result in limited domain understanding and reduced factual precision when encountering specialized terminology, policies, or metrics.

Second, the interpretation of visual artifacts (e.g., histograms, scatter plots) is highly sensitive to prompt quality and input clarity. When charts contain incomplete labels or ambiguous visual encodings, the system may produce inaccurate or overly generic summaries. This remains a key challenge for applying large language models to visual analytics where precise data interpretation is required for decision support.

Third, the current architecture does not perform global consistency checks across narrative stages. Consequently, minor inaccuracies introduced during early narration may propagate through later stages if not corrected by the LLM-as-a-judge agent. While human oversight mitigates this risk, achieving reliable full automation will require additional safeguards to maintain factual consistency and stability.

Finally, inference scalability and agent coordination overhead present further constraints. Reported execution time and cost metrics account for both model inference and inter-agent communication. Because the current prototype operates sequentially, coordination overhead arises mainly from message exchange among narration, evaluation, and synthesis agents. Although modest within the evaluated dataset, larger applications may require distributed or asynchronous execution. Future research should therefore investigate scalable orchestration and parallel processing strategies to improve computational efficiency and reproducibility in complex data environments.

*6.4. Domain-Specific Challenges, Generalizability, and Ethical Implications*

6.4.1. Generalizability and Practical Adaptation

The proposed LLM-based multi-agent framework introduces several practical and domain-specific considerations. A key limitation of this study is its validation on a single real-world case: fuel efficiency analysis in public transportation. Although this provides a representative and data-rich context, further demonstrations across other energy-intensive and industrial domains, such as manufacturing analytics, process optimization, or fault diagnosis [39,40], are needed to assess the framework's broader applicability. These cross-domain applications also reveal ongoing challenges in domain adaptation and factual reliability, which the proposed framework addresses through evaluation-driven coordination and human oversight [39,40]. The modular architecture, which separates data narration, evaluation, and synthesis into independent agents, is intentionally domain-agnostic. By substituting analytical artifacts such as sensor data, production metrics, or vibration signals, the same workflow can be adapted to various operational contexts. Extending future work to these domains would provide stronger empirical validation and demonstrate the framework's potential to enhance automated reporting and decision support across heterogeneous industrial systems.

6.4.2. Ethical and Human-in-the-Loop Considerations

Ethical and human oversight aspects are equally significant for the responsible use of AI-driven frameworks in energy management. Automated narrative generation can influence operational and policy decisions, requiring transparency, accountability, and explainability. The proposed architecture promotes traceability through structured data-to-narrative workflows and retains human oversight via an expert-in-the-loop evaluator. This ensures that automation complements rather than replaces human expertise. Data integrity and fairness also remain critical. While the current framework uses aggregated, non-personal data, future applications involving industrial monitoring or human–machine interactions must adopt privacy safeguards and bias mitigation measures. Ethical deployment further depends on transparent documentation, reproducibility of results, and mechanisms for expert feedback. Collaborative governance among engineers, domain experts, and decision-makers is essential to maintain accountability and trust in AI-assisted analytics.

*6.5. Future Research Directions*

Future work should address these challenges through targeted enhancements. Fine-tuning LLMs on domain-specific corpora, including transport energy datasets, policy documents, and technical standards, could improve contextual accuracy and reduce hallucinations. Incorporating retrieval-augmented generation (RAG) techniques or structured knowledge bases would allow the system to ground narratives in verifiable sources, improving factual reliability and traceability.

To mitigate error propagation, future iterations of the framework could implement cross-agent validation mechanisms and uncertainty estimation modules. These would enable agents to identify inconsistent outputs or low-confidence predictions and trigger corrective steps or human review. Explainability features, such as rationale generation or evidence linking, would further enhance trust and usability.

Scalability across different domains, such as building energy management or industrial energy optimization, will require modular extensions to handle diverse data formats and stakeholder expectations. A multi-domain evaluation campaign, supported by a benchmark dataset for multimodal narration in energy informatics, would be a critical step toward generalizability and standardization.

Lastly, Future research should extend the proposed framework to additional domains to evaluate its scalability and robustness. Further work is also needed to formalize ethical governance mechanisms and strengthen human-in-the-loop validation, ensuring that AI-assisted narration and decision support remain transparent, accountable, and aligned with expert oversight in complex energy and industrial environments.

## 7. Conclusions

This study proposed a novel multi-agent framework that leverages multimodal Large Language Models (LLMs) to automate the generation of interpretive narratives and decision-oriented insights in energy informatics workflows. Designed to address the communication gap between technical analytics and operational decision-making, the framework integrates dedicated agents for data narration, evaluation, and expert oversight, enabling a structured, iterative approach to translating complex outputs into accessible, stakeholder-relevant reports.

The research advances the field by operationalizing LLM-based insight generation in the context of public transport energy efficiency, demonstrating how AI technologies can support meaningful applications in smart mobility systems. Unlike conventional approaches that focus narrowly on visualization or narration, the framework emphasizes process integration, quality assurance, and scalability, providing a practical pathway for deploying large language models in data-rich, energy-critical domains.

Beyond its application to public transportation, the proposed framework is designed to be generalizable across diverse industrial and energy domains, including manufacturing analytics, process monitoring, and fault diagnosis. Its modular architecture allows straightforward adaptation to new data modalities and decision contexts. Furthermore, the incorporation of human-in-the-loop evaluation supports responsible and transparent deployment, ensuring that AI-generated insights remain explainable, verifiable, and aligned with expert judgment. Together, these features position the framework as a scalable and ethically robust foundation for AI-assisted decision support in energy and industrial systems.

administration, B.N.J. and Z.G.M.; funding acquisition, B.N.J. All authors have read and agreed to the published version of the manuscript.

**Funding:** This work is funded by the Energy Technology Development and Demonstration Programme (EUDP) in Denmark under the project EUDP 2021-II Driver Coach [case no. 64021-2034].

**Institutional Review Board Statement:** Not applicable.

**Informed Consent Statement:** Not applicable.

**Data Availability Statement:** The raw data supporting the conclusions of this article will be made available by the authors on request.

**Conflicts of Interest:** The authors declare no conflicts of interest.

## Abbreviations

The following abbreviations are used in this manuscript:

| | |
|---|---|
| AI | Artificial Intelligence |
| ANID | Accurate Narrative Information Density |
| API | Application Programming Interface |
| ARI | Automated Readability Index |
| CCoT | Contrastive Chain-of-Thought |
| CoT | Chain-of-Thought |
| DN | DataNarrative (prompting strategy) |
| DS-Agent | Data Science Agent |
| FKGL | Flesch-Kincaid Grade Level |
| FRE | Flesch Reading Ease |
| GMM | Gaussian Mixture Model |
| IT | Information Technology |
| LLM | Large Language Model |
| NRDS | Number of Recommendations with Data Support |
| PDE | Probability Density Estimate |
| PRDS | Percentage of Recommendations with Data Support |
| RAG | Retrieval-Augmented Generation |
| RL | Report Length |
| RLHF | Reinforcement Learning with Human Feedback |
| SHAP | SHapley Additive exPlanations |
| SQL | Structured Query Language |
| USD | United States Dollar |

## Appendix A. Prompt Engineering

This section presents the key prompts and the associated augmented files used throughout the data narration process. Listing A1 illustrates the prompt designed for generating data narratives. Listing A2 depicts the prompt utilized by the LLM-as-a-judge agent. Listing A3 shows the prompt applied during the integration stage for report generation. Listing A4 outlines the evaluation criteria employed within the LLM-as-a-judge agent. Finally, Listing A5 provides the template used for multimodal data narration.

**Listing A1.** The prompt used to generate data narratives.

[System Prompt]

You are a genius data analyst who can process and explain multimodal scientific data. You have domain specific knowledge in transportation and fuel efficiency. Firstly, I will give you a background file, introducing the project background and raw data. Next, I will give you an image, which can be a chart or another type of data representation. Please describe and interpret the context of this image.

[User Prompt]

Here is the background file: {background}

Firstly, based on the content in the file and your global knowledge, please generate the domain knowledge in this field and store this knowledge in your mind.

Secondly, based on your generated domain knowledge and the background file, please think step by step to solve the following task.

The task is as follows:

1. For the uploaded image, please describe the overall scene and the main objects present in the uploaded image.
2. Interpret the image's context and explain what story or message it might be conveying.

The output should follow the template in the file: {template}

**Listing A2.** The prompt used to evaluate data narratives.

[System Prompt]

You are a genius and experienced data analyst who can process text files and write high-quality report. The report is aimed to make recommendations and decision support for the stakeholders in bus companies without technical background.

[User Prompt]

Based on the project introduction in the file: {background}, and your global knowledge, please generate the global domain knowledge in this field and store the knowledge in your mind. Now, please read and analyze the following files: ..., which store the analysis and explanation of scientific figures related to this project.

The project description and motivation are in {background}.

Please write a high-quality report to make recommendations for the stakeholders in bus companies.

The report should follow the following tasks:

1. It includes the data description, the clustering analysis, and the post hoc analysis including driver distribution, route distribution and route type distribution.
2. Necessary statistical data should be involved in each part of the report to support the analysis and recommendation.
3. Do not make up the fake statistical values.
4. The report should be in the form of an article, not bullet points, and must be at least 600 words.

Before generating the report, please think it over step by step, but do not output your thinking steps.

**Listing A3.** The prompt used to generate the report.

[System Prompt]

You are a genius and fair data analyst and evaluator that can process and assess text files generated by another LLM agent accurately. I will give you two files. The first one is the project introduction. The second is the generated context waiting for your evaluation. Please return the results as the JSON output.

[User Prompt]

Based on the project introduction in the file: {background}, and your global knowledge, please generate the global domain knowledge in this field and store the knowledge in your mind. Next, based on the project introduction, please evaluate the file {text}, generated by another LLM agent, which explains an image of data analytics results in this project following the following steps:

1. Understand the assessment criteria, including Clarity, Relevance, Insightfulness, and Contextualization. Their definition and score scale are introduced in {eva_criteria}.

2. Evaluate the file based on each defined criterion.
3. Provide score for each criterion and explain the reason.
4. Conclude the overall evaluation results and provide a summary.

Please return your result in this exact JSON format:

{"evaluation_scores": {

"Clarity": <score>,

"Relevance": <score>,

"Insightfulness": <score>,

"Contextualization": <score>,

"Overall_score": <average score of above four>,

},

"evaluation_report": <providing scores of each criterion, explanation of scores of each criterion, overall evaluation results and summary>}

**Listing A4.** Evaluation criteria in the LLM-as-a-judge agent.

**1. Clarity:** Clarity refers to how well the descriptions convey the message without ambiguity. Information should be presented logically and in a way that is easily understood by the intended audience.

- **0—Inadequate**: Information is confusing and unclear.
- **1—Needs Improvement**: Some parts are clear, but overall understanding is hindered.
- **2—Satisfactory**: Generally clear but lacks depth or contains minor ambiguities.
- **3—Proficient**: Very clear, easy to understand, logical flow, with minimal effort needed for comprehension.
- **4—Excellent**: Exceptionally clear; no ambiguity, and very engaging in presentation.

**2. Relevance:** Relevance assesses whether the content directly pertains to the project objectives and the data being analyzed.

- **0—Inadequate**: Content is largely irrelevant and does not relate to the project objectives.
- **1—Needs Improvement**: Some relevant points, but many unrelated aspects.
- **2—Satisfactory**: Mostly relevant, with some minor digressions.
- **3—Proficient**: Highly relevant, with all points supporting project objectives.
- **4—Excellent**: Entirely relevant; addresses the core analytic objectives with precision.

**3. Insightfulness:** Insightfulness measures how well the content provides unique or valuable insights derived from the data.

- **0—Inadequate**: No insights are provided; merely descriptive.
- **1—Needs Improvement**: Some insights, but largely superficial or expected.
- **2—Satisfactory**: Provides some valuable insights but lacks depth.
- **3—Proficient**: Strong insights that provoke thought or highlight significant implications.
- **4—Excellent**: Deep insights that provide new avenues of thought.

**4. Contextualization:** This evaluates how well the data and conclusions are placed within a broader context, such as environmental or policy implications.

- **0—Inadequate**: No contextual information provided.
- **1—Needs Improvement**: Minimal context; does not fully connect findings to the broader implications.
- **2—Satisfactory**: Some contextual elements present, but could be enhanced for better clarity.
- **3—Proficient**: Good contextualization, connecting findings to relevant implications.
- **4—Excellent**: Comprehensive context that frames the findings within larger environmental and policy discussions.

**Listing A5.** Template for data narration.

1. [Chart] Type and Purpose

This is a [type of chart] showing [very brief statement of what is being compared or analyzed].

2. Key Variables and Metrics

If it is a chart, the output should be:

The $x$-axis denotes [list main variables/metrics] with [units or scale if applicable].

The $y$-axis denotes [list main variables/metrics] with [units or scale if applicable].

3. Main Findings and Trends

The primary observation is [dominant pattern or trend].

[Mention any secondary or more observations if needed].

The observed trend shows [the insights and explanations of trends and outliers].

4. Statistical Insights

The [type of chart] indicates [basic statistical result, e.g., $p$-value, $R^2$, mean, etc.].

The [type of chart] illustrates [insights and analysis of interconnected attributes].

5. Contextual Implications

This suggests [main relevance or application in industrial/data science context].

## Appendix B. Examples

This section presents typical examples of data narrations generated during the experiments on model and prompt selection, as well as ablation studies. Table A1 shows examples from the model and prompt selection experiments, while Table A2 presents examples from the ablation studies.

**Table A1.** Examples of experiments of model and prompt selection.

| No. | Narratives | Model and Prompting Method | Comments |
|---|---|---|---|
| 1 | This is a histogram showing the distribution of fuel efficiency values measured in liters per 100 km (L/100 km). | CoT + GPT-4.1 mini | N/A |
| | This is a histogram showing the distribution of fuel efficiency for a dataset of vehicles. | CoT + Gemini 2.5 flash | It is short and misses some information. |
| 2 | The $y$-axis denotes the number of occurrences or frequency of vehicles within each fuel efficiency range, with values ranging from 0 to 2500. | CoT + GPT-4.1 mini | N/A |
| | The $y$-axis denotes the number of observations or frequency. | CoT + Llama 4 maverick | It does not describe a specific chart. |
| 3 | This is a ternary scatter plot showing the distribution and clustering of data points based on three variables: Rural, Urban, and Highway. | CoT + GPT-4.1 mini | N/A |
| | This is a triangular scatter plot showing the distribution of data points across three categorical dimensions: Urban, Rural, and Highway. | CoT + Claude 3.5 haiku | "triangular scatter plot" is not a formal and accurate name. |
| 4 | The distribution illustrates that while some routes have low entropy, indicating predictability, others have moderate to high entropy, reflecting more variability in route types within the cluster. | CoT + GPT-4.1 mini | It is detailed and has better explanations. |
| | The histogram illustrates the distribution shape and highlights multiple common entropy levels rather than a uniform spread. | CCoT + GPT-4.1 mini | This explanation focuses on statistics. |

| | The distribution is bimodal, suggesting two dominant patterns of route type diversity within the cluster. | DN + GPT-4.1 mini | It is incorrect, there are three dominant patterns. |

**Table A2.** Examples of the ablation study.

| No. | Narratives | Model and Prompting Method | Comments |
| --- | --- | --- | --- |
| 1 | This is a density plot combined with a scatter plot showing the distribution and clustering of fuel efficiency values for bus trips in public transportation. | CoT+B/baseline | "CoT+B" and baseline are same in this sentence. "bus trips in public transportation" is generated from background. |
| | This is a density plot combined with a scatter plot showing the distribution and clustering of fuel efficiency values (measured in liters per 100 km) across different clusters. | CoT | No background information. |
| 2 | The primary observation is that a large proportion of drivers have very low entropy values close to 0, indicating that these drivers tend to have bus trips concentrated in a single cluster. | CoT+B/baseline | "CoT+B" and baseline are same in this sentence. More concise. |
| | The primary observation is that a large proportion of drivers have very low entropy values close to 0, indicating that most drivers tend to have bus trips concentrated in a single cluster rather than spread evenly across all four clusters. | CoT+B+E | Longer but do not provide more useful information. |
| 3 | The initial Cluster 0 is divided into two distinct clusters: Cluster 0 and Cluster 3. Cluster 0 shows a wider spread with lower density at moderate fuel efficiency values, while Cluster 3 appears at higher fuel efficiency values (above 80 L/100 km). | Baseline | This is provided through human intervention, which cannot be generated through data reasoning. |

## Appendix C. Report

This section presents the technical report generated by the framework, as illustrated in Listing A6. Due to the report's length, the results primarily highlight its overall structure and organization.

**Listing A6.** Generated technical report.

---

**Report on Fuel Efficiency Analysis in Public Bus Transportation: Insights and Recommendations for Stakeholders**

**Introduction**

This report presents a comprehensive analysis of fuel efficiency in public bus transportation based on data collected from 4006 bus trips …, during the week of ... The primary objective of the study is to … Employing Gaussian Mixture Model (GMM) clustering, … Subsequent post hoc analyses explore …, providing actionable insights for stakeholders …

**Data Description**

The dataset comprises fuel efficiency measurements expressed in liters per 100 km (L/100 km) for each bus trip, defined as … The distribution of fuel efficiency values reveals that …. Notably, the distribution is right-skewed, with a long tail extending towards … This pattern indicates that …, suggesting variability in ...

**Clustering Analysis**

The application of GMM clustering effectively partitions the bus trips into four distinct clusters, …. Cluster 0, ... Cluster 1, the largest cluster encompassing 2857 trips, centers around moderate fuel efficiency with a mean of 46.03 L/100 km and a range from 29.16 to 56.02 L/100 km. Cluster 2 …, while Cluster 3, …

The density plot and scatter visualization confirm that …

**Post Hoc Analysis: Driver Distribution**

Examining the distribution of clusters across individual drivers reveals that most drivers predominantly operate within Cluster 1, indicating ... Entropy analysis of driver trip distributions further supports these findings …

**Post Hoc Analysis: Route Distribution**

The distribution of clusters across bus routes shows that … Entropy analysis of routes reveals that … These insights imply that …

**Post Hoc Analysis: Route Type Distribution**

The analysis of route type composition across clusters highlights ... Cluster 2 trips show moderate route type diversity, with a tendency towards …

**Recommendations**

Based on the comprehensive analysis, several recommendations emerge for stakeholders in bus companies aiming to enhance fuel efficiency:

1. Driver-Focused Interventions: …
2. Route-Specific Strategies: …
3. Route Type Considerations: …
4. Monitoring and Continuous Analysis: …

**Conclusion**

This study demonstrates that fuel efficiency in public bus transportation is influenced by ... The use of GMM clustering reveals ... By leveraging these insights, bus companies can …

25 of 25